\documentclass[letterpaper]{article}
\usepackage{proceed2e}
\usepackage[margin=1in]{geometry}
\usepackage{graphicx}
\usepackage{booktabs}
\usepackage{multirow}
\usepackage{algorithm}% http://ctan.org/pkg/algorithms
\usepackage[noend]{algpseudocode}% http://ctan.org/pkg/algorithmicx

\usepackage{amsmath}
\usepackage{amssymb}
\usepackage{bm}

\usepackage{times}

\newcommand{\X}{\ensuremath{\mathbf{X}}}
\newcommand{\Y}{\ensuremath{\mathbf{Y}}}
\newcommand{\x}{\ensuremath{\mathbf{x}}}
\newcommand{\y}{\ensuremath{\mathbf{y}}}

\newcommand{\region}{\ensuremath{\mathbf{R}}}
\newcommand{\partition}{\ensuremath{\mathcal{P}}}
\newcommand{\rg}{\ensuremath{\mathcal{R}}}

\newcommand{\SPN}{\mathcal{S}}

\newcommand{\Node}{\mathsf{N}}
\newcommand{\SumNode}{\mathsf{S}}
\newcommand{\ProdNode}{\mathsf{P}}

\newcommand{\Nodes}{\bm{\mathsf{N}}}

\newcommand{\ch}{\mathbf{ch}}
\newcommand{\scope}{\mathbf{sc}}

\newcommand{\w}{w}

\usepackage{xcolor}

\title{Probabilistic Deep Learning using Random Sum-Product Networks}

\author{
{\bf Robert Peharz} \\ 
 Dept. of Engineering \\ 
 University of Cambridge \\
\mbox{} \\
{\bf Martin Trapp} \\
 Austrian Research Institute \\
 for Artificial Intelligence \\
\And
{\bf Antonio Vergari} \\
 Max Planck Institute \\
 for Intelligent Systems \\
\mbox{} \\
{\bf Kristian Kersting} \\
 Computer Science Dept. \\
 TU Darmstadt \\
\And 
{\bf Karl Stelzner} \\
 Computer Science Dept. \\
 TU Darmstadt \\
\mbox{} \\
{\bf Zoubin Ghahramani} \\
  Dept. of Engineering \\ 
 University of Cambridge \\
\And 
{\bf Alejandro Molina} \\
 Computer Science Dept. \\
 TU Darmstadt \\
}

\begin{document}

\maketitle

\begin{abstract}
The need for consistent treatment of uncertainty has recently triggered increased interest in probabilistic deep learning methods.
However, most current approaches have severe limitations when it comes to inference, since many of these models do not even permit to evaluate exact data likelihoods.
Sum-product networks (SPNs), on the other hand, are an excellent architecture in that regard, as they allow to efficiently evaluate likelihoods, as well as arbitrary marginalization and conditioning tasks.
Nevertheless, SPNs have not been fully explored as serious deep learning models, likely due to their special structural requirements, which complicate learning.
In this paper, we make a drastic simplification and use random SPN structures which are trained in a ``classical deep learning manner", i.e.~employing automatic differentiation, SGD, and GPU support.
The resulting models, called RAT-SPNs, yield prediction results comparable to deep neural networks, while
still being interpretable as generative model and maintaining well-calibrated uncertainties.
This property makes them highly robust under missing input features and enables them to naturally detect outliers and peculiar samples.
\end{abstract}

%%%%%%%%%%%%%%%%%%%%%%%%%%%%%%%%%%%%%%%%%%%%%%%%%%%%%%%%%%%%%%%%%%%%%%%%%%%%%%%%%%%%%%1
%%%%%%%%%%%%%%%%%%%%%%%%%%%%%%%%%%%%%%%%%%%%%%%%%%%%%%%%%%%%%%%%%%%%%%%%%%%%%%%%%%%%%%
%%%%%%%%%%%%%%%%%%%%%%%%%%%%%%%%%%%%%%%%%%%%%%%%%%%%%%%%%%%%%%%%%%%%%%%%%%%%%%%%%%%%%%
\section{INTRODUCTION}

Uncertainty matters.
An intelligent system applied in the real world should both be able to deal with uncertain inputs, as well as express its uncertainty over outputs. 
Especially the latter is a crucial point in automatic decision-making processes, such as medical diagnosis and planning systems for autonomous agents. 
Therefore, it is no surprise that probabilistic approaches have recently gained great momentum in deep learning, having led to a variety of innovative probabilistic models such as variational autoencoders (VAEs) \cite{Kingma2014}, deep generative models \cite{Rezende2014}, generative adversarial nets (GANs) \cite{Goodfellow2014}, neural auto-regressive density estimators (NADEs) \cite{Larochelle2011}, and Pixel-CNNs/RNNs \cite{vandenOord2016}.
However, most of these probabilistic deep learning systems have limited capabilities when it comes to \emph{inference}.

Implicit probabilistic models like GANs, even when successful in capturing the data distribution, do not allow to evaluate the likelihood of a test sample.
Similar problems arise in deep generative models and VAEs, which typically employ a jointly trained inference network to infer the posterior over a latent variable space.
While these techniques mark a milestone in variational learning, inference in these models and -- ironically -- also their inference networks is limited to drawing samples, forcing users to retreat to Monte Carlo estimates.
Auto-regressive density estimators like NADEs and Pixel-CNNs/RNNs somewhat alleviate these limitations, as they permit exact and efficient evaluation of model likelihoods.
Moreover, they permit certain marginalization and conditioning tasks, as long as the task is consistent with the variable ordering assumed in the model.
Inference tasks not consistent with this variables ordering, however, remain intractable.
Uria et al.~\cite{Uria2014} address this problem by training an ensemble of NADEs with shared network structure. 
This approach, however, introduces the delicate problem of approximately training a super-exponential ensemble of NADEs. 
Thus, auto-regressive models still fall short when fully fledged inference is required.

To this end, \emph{sum-product networks} (SPNs) \cite{Poon2011} are a promising remedy, as they are a class of probabilistic model which permits \emph{exact} and \emph{efficient} inference.
More precisely, SPNs are able to compute any marginalization and conditioning query in time linear of the model's representation size.
Nevertheless, although SPNs can be described in a nutshell as ``deep mixture models'' \cite{Peharz2016}, they have received rather limited attention in the deep learning community, despite their attractive inference properties.
We conjecture that there are three reasons why SPNs have been under-used in deep learning so far.

First, the structure of SPNs needs to obey certain constraints, requiring either careful structure design by hand or learning the structure from data \cite{Dennis2012,Gens2013,Peharz2013,Rooshenas2014,Vergari2015,Adel2015,Trapp2016}.
These structural requirements are somewhat opposed to the usual homogeneous model structures employed in deep learning, i.e.~combining modules like matrix multiplication and element-wise non-linearities in an almost unconstrained way.
Second, the parameter learning schemes proposed so far are usually inspired by graphical models \cite{Poon2011,Zhao2016,Peharz2016} or tailored to SPNs \cite{Gens2012}.
This peculiar learning style probably hindered a wide application of SPNs in the connectionist approach so far, which typically relies on automatic differentiation and SGD.
Third, there seems to be a folklore that SPNs are ``somewhat weak function approximators'', i.e.~it is widely believed that SPNs are significantly inferior to solve prediction tasks to an extent we expect from deep neural networks.
Indeed, in \cite{Martens2015}, a class of distributions was identified which can be tractably represented by a neural net, but not by an SPN.
However, as also mentioned in \cite{Martens2015}, this example is a somewhat academic one, and we should not jump to conclusion concerning SPNs' fitness in practical problems.
Furthermore, the notion of SPNs used here was a restricted one, i.e.~using uni-variate leaves, and the example could actually be circumvented by extending SPNs to multi-variate leaves \cite{Peharz2015b}.
In that way, the tractable representation of the negtive example in \cite{Martens2015} could (trivially) by incorporated in the SPN framework.
In general, SPNs inherit universal approximation properties from mixture models, as a mixture model is simply a ``shallow'' SPN with a single sum node.
Consequently, SPNs are able to approximate any prediction function via probabilistic inference in an asymptotic sense.

In this paper, we investigate the fitness of SPNs as deep learning models from a practical point of view.
To this end, we introduce a particularly simple way to construct SPNs, waiving the necessity for structure learning and simplifying their use as connectionist model.
These SPNs are obtained by first constructing a random \emph{region graph} \cite{Dennis2012,Peharz2013} laying out the overall network structure, and subsequently populating the region graph with tensors of SPN nodes.
This architecture -- which we call \emph{Random Tensorized SPNs} (RAT-SPNs) -- is naturally implemented in deep learning frameworks like as TensorFlow \cite{Abadi2015} and easily optimized end-to-end using automatic differentiation, SGD, and automatic GPU-parallelization.
To avoid overfitting, we adopt the well-known dropout heuristic \cite{Srivastava2014}, which yields an elegant probabilistic interpretation in our models as marginalization of missing features (dropout at inputs) and as injection of discrete noise (dropout at sum nodes).
We trained RAT-SPNs on several real-world classification data sets, showing that their prediction performances are comparable to traditional deep neural networks.
At the same time, RAT-SPNs maintain a complete joint distribution over both inputs and outputs, which allows us to treat uncertainty in a consistent manner.
We show that RAT-SPNs are dramatically more robust in the presence of missing features than neural networks.
Furthermore, we demonstrate that RAT-SPNs also provide well-calibrated uncertainty estimates over their inputs, i.e., the model ``knows what it does not know''.
This property can be naturally exploited for anomaly and out-of-domain detection.

The paper is organized as follows.
Section~\ref{sec:background} reviews SPNs and required background.
In section~\ref{sec:rat_spns}, we introduce RAT-SPNs and discuss implementation and training. 
Experimental results are presented in section \ref{sec:experiments} and section~\ref{sec:conclusion} concludes the paper.

%%%%%%%%%%%%%%%%%%%%%%%%%%%%%%%%%%%%%%%%%%%%%%%%%%%%%%%%%%%%%%%%%%%%%%%%%%%%%%%%%%%%%%
%%%%%%%%%%%%%%%%%%%%%%%%%%%%%%%%%%%%%%%%%%%%%%%%%%%%%%%%%%%%%%%%%%%%%%%%%%%%%%%%%%%%%%
%%%%%%%%%%%%%%%%%%%%%%%%%%%%%%%%%%%%%%%%%%%%%%%%%%%%%%%%%%%%%%%%%%%%%%%%%%%%%%%%%%%%%%
\section{RELATED WORK}   \label{sec:background}

We denote random variables (RVs) by upper-case letters, e.g.~$X$, $Y$, and their values by corresponding lower-case letters, e.g.~$x$, $y$.
Similarly, we denote sets of RVs by upper-case boldface letters, e.g.~$\X$, $\Y$ and their combined values by corresponding lower-case letters, e.g.~$\x$, $\y$.
An SPN $\SPN$ over $\X$ is a probabilistic model defined via a directed acyclic graph (DAG) containing three types of nodes: 
\emph{input distributions}, \emph{sums} and \emph{products}.
All leaves of the SPN are distribution functions over some subset $\Y \subseteq \X$.
Inner nodes are either weighted sums or products, denoted by $\SumNode$ and $\ProdNode$, respectively, i.e., $\SumNode = \sum_{\Node \in \ch(\SumNode)} \w_{\SumNode, \Node} \Node$ and $\ProdNode = \prod_{\Node \in \ch(\ProdNode)} \Node$, where $\ch(\Node)$ denotes the children of $\Node$.
The sum weights $\w_{\SumNode,\Node}$ are assumed to be non-negative and normalized, i.e., $\w_{\SumNode,\Node} \geq 0$, $\sum_\Node \w_{\SumNode,\Node} = 1$.

The \emph{scope} of an input distribution $\Node$ is defined as the set of RVs $\Y$ for which $\Node$ is a distribution function, i.e.~$\scope(\Node): = \Y$.
The scope of an inner (sum or product) node $\Node$ is recursively defined as 
$\scope(\Node) = \bigcup_{\Node' \in \ch(\Node)} \scope(\Node')$.
To allow efficient inference, SPNs should satisfy two structure constraints \cite{Darwiche2003,Poon2011}, namely \emph{completeness} and \emph{decomposability}.
An SPN is complete if for each sum $\SumNode$ it holds that $\scope(\Node') = \scope(\Node'')$, for all $\Node', \Node'' \in \ch(\SumNode)$.
An SPN is decomposable if it holds for each product $\ProdNode$ that $\scope(\Node') \cap \scope(\Node'') = \emptyset$, for all $\Node' \not= \Node'' \in \ch(\ProdNode)$.
In that way, all nodes in an SPN recursively define a distribution over their respective scopes:
the leaves are distributions by definition, sum nodes are mixtures of their child distributions, and products are factorized distributions, i.e., assuming (conditional) independence among the scopes of their children.

Besides \emph{representing} probability distributions, the crucial advantage of SPNs is that they allow \emph{efficient inference}: In particular, any marginalization task reduces to the corresponding marginalizations at the leaves (each leaf marginalizing only over its scope), and evaluating the internal nodes in a bottom-up pass \cite{Peharz2015b}.
Thus, marginalization in SPNs follows essentially the same procedure as evaluating the likelihood of a sample -- both scale linearly in the SPN's representation size.
Conditioning is tackled in a similar manner.

Learning the parameters of SPNs, i.e.~the sum weights and the parameters of input distributions, can be addressed in various ways.
By interpreting the sum nodes as discrete latent variables \cite{Poon2011,Zhao2015,Peharz2016}, SPNs can be trained using the classical expectation-maximization (EM) algorithm \cite{Peharz2016}.
Zhao et al.~\cite{Zhao2016b} derived a concave-convex procedure, which interestingly coincides with the EM updates for sum-weights.
Moreover, SPN parameters can be treated in the Bayesian framework, as proposed in \cite{Rashwan2016,Zhao2016,Trapp2016}.
Trapp et al.~\cite{Trapp2017} introduced a safe semi-supervised learning scheme for discriminative and generative parameter learning, providing guarantees for the performance in the semi-supervised case. 
Vergari et al.~\cite{Vergari2018} employed SPNs as probabilistic autoencoders and unsupervised representation learners.
The structure of SPNs can be crafted by hand \cite{Poon2011,Peharz2014a} or learned from data.
Most structure learners \cite{Rooshenas2014,Vergari2015,Adel2015,Molina2018} are variations of the divide-and-conquer scheme known as \emph{LearnSPN} \cite{Gens2013}. 
This schemed recursively splits the data via clustering (determining sum nodes) and independence tests (determining product nodes).

In general, most approaches to learning SPNs are motivated by techniques borrowed from the graphical model literature.
However, from their definition it is evident that SPNs can also be interpreted as a special kind of neural networks.
In this paper, we aim to follow through this interpretation and investigate how SPNs perform when treated as connectionist model.
To this end, we make a drastic simplification by simply picking a scalable random structure and optimizing its parameters in a deep learning manner.

%%%%%%%%%%%%%%%%%%%%%%%%%%%%%%%%%%%%%%%%%%%%%%%%%%%%%%%%%%%%%%%%%%%%%%%%%%%%%%%%%%%%%%
%%%%%%%%%%%%%%%%%%%%%%%%%%%%%%%%%%%%%%%%%%%%%%%%%%%%%%%%%%%%%%%%%%%%%%%%%%%%%%%%%%%%%%
%%%%%%%%%%%%%%%%%%%%%%%%%%%%%%%%%%%%%%%%%%%%%%%%%%%%%%%%%%%%%%%%%%%%%%%%%%%%%%%%%%%%%%
\section{RANDOM TENSORIZED SUM-PRODUCT NETWORKS}   \label{sec:rat_spns}

\begin{figure*}
\begin{centering}
\includegraphics[width=0.9\textwidth]{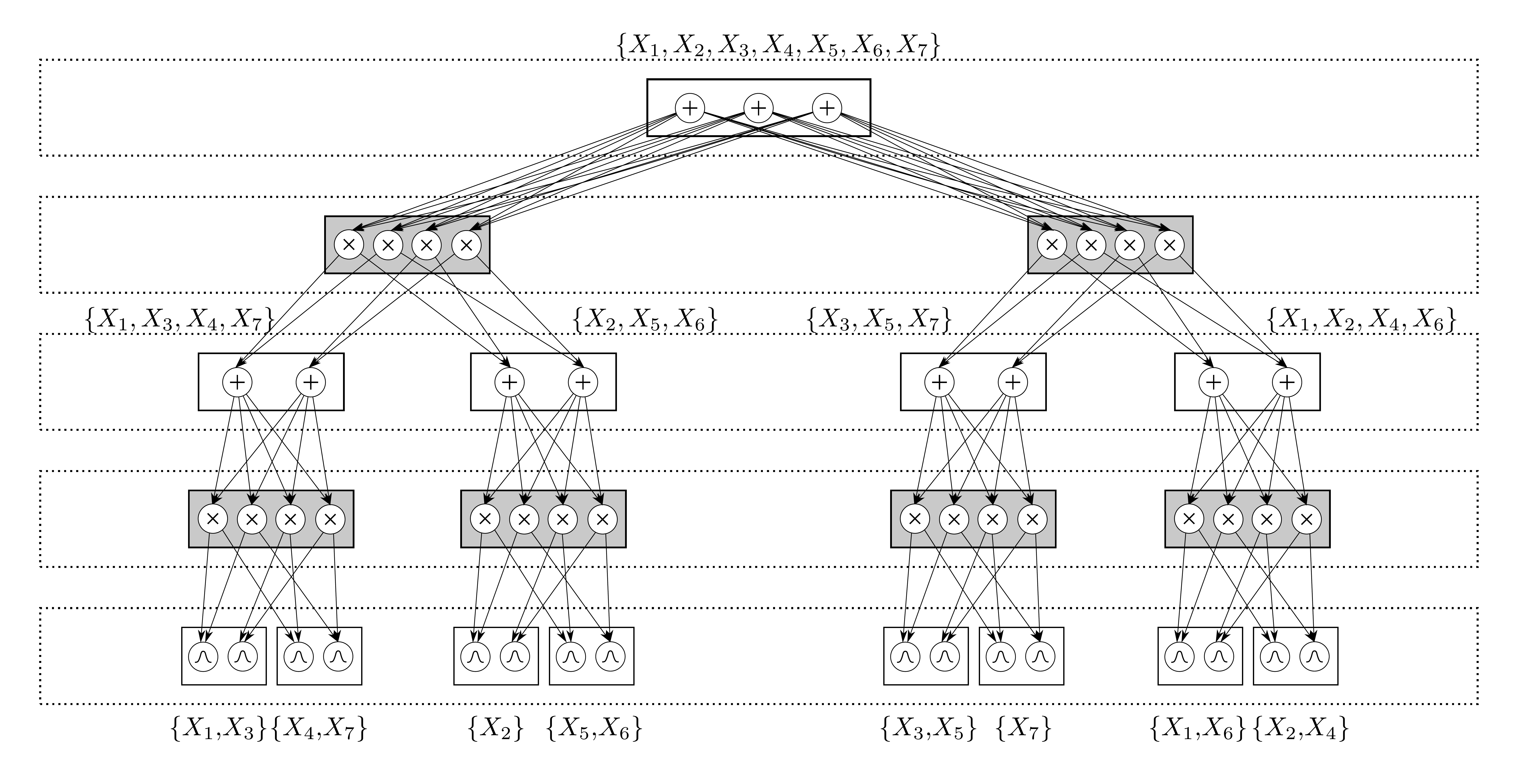}
\caption{Example RAT-SPN over 7 RVs $\{X_1 \dots X_7\}$, using parameters $C=3$, $D=2$, $R=2$, $S=2$, and $I=2$ in Algorithm~\ref{alg:construct_random_region_graph} and Algorithm~\ref{alg:construct_spn_from_region_graph}.
}   
\label{fig:example_RAT-SPN}
\end{centering}
\end{figure*}

In order to construct \emph{random-and-tensorized SPNs} (RAT-SPNs) we use a \emph{region graph} \cite{Poon2011,Dennis2012,Peharz2013} as an abstract representation of the network structure.
Given a set of RVs $\X$, a \emph{region} $\region$ is defined as any non-empty subset of $\X$.
Given any region $\region$, a $K$-\emph{partition} $\partition$ of $\region$ is a collection of $K$ non-empty, non-overlapping  subsets $\region_1, \dots, \region_K$ of $\region$, whose union is again $\region$, i.e., $\partition = \{\region_1, \dots, \region_K\}$, $\forall k \colon \region_k \not= \emptyset$, $\forall k \not= l \colon \region_k \cap \region_l = \emptyset$, $\bigcup_k \region_k = \X$. 
In this paper, we consider only 2-partitions, which causes all product nodes in our SPNs to have exactly two children.
This assumption, frequently made in the SPN literature, simplifies SPN design and seems not to impair performance.

A \emph{region graph} $\rg$ over $\X$ is a DAG whose nodes are regions and partitions such that 
i) $\X$ is a region in $\rg$ and has no parents (\emph{root region}), 
ii) all other regions have at least one parent,
iii) all children of regions are partitions and all children of partitions are regions (i.e., $\rg$ is bipartite),
iv) if $\partition$ is a child of $\region$, then $\bigcup_{\region' \in \partition} \region' = \region$ and
v) if $\region$ is a child of $\partition$, then $\region \in \partition$.
From this definition it follows that a region graph dictates a hierarchical partition of the overall scope $\X$.
We denote regions which have no child partitions as \emph{leaf regions}.

Given a region graph, we can construct a corresponding SPN, as illustrated in Algorithm~\ref{alg:construct_spn_from_region_graph}.
In this paper we assume a classification problem with $C$ classes (for density estimation we simply set $C=1$), where each class conditional distributions corresponds to a root of the RAT-SPN, i.e.~the $k^\text{th}$ root represents $p(\X \, | \, C=k)$.
By multiplying the SPN roots with a prior $p(C)$, we get a full joint distribution $p(\X, C)$.
Further, $I$ is the number of input distributions per leaf region, and $S$ is the number of sum nodes in regions, which are neither leaf nor root regions.
It is easy to verify that Algorithm~\ref{alg:construct_spn_from_region_graph} always leads to a complete and decomposable SPN.
\begin{algorithm}[t]
\caption{Construct SPN from Region Graph}   \label{alg:construct_spn_from_region_graph}
\begin{algorithmic}[1]
\Procedure{ConstructSPN}{$\rg,C,S,I$}
\State Make empty SPN
\For{$\region \in \rg$} 
\If{$\region$ is a leaf region}
\State Equip $\region$ with $I$ distribution nodes
% \ElsIf{$\region = \X$}
\ElsIf{$\region$ is the root region}
\State Equip $\region$ with $C$ sum nodes
\Else
\State Equip $\region$ with $S$ sum nodes
\EndIf
\EndFor
\For{$\partition = \{\region_1, \region_2\} \in \rg$} 
\State Let $\Nodes_\region$ be the nodes for region $\region$
\For{$\Node_1 \in \Nodes_{\region_1}, \Node_2 \in \Nodes_{\region_2}$}
\State Introduce product $\ProdNode = \Node_1 \times \Node_2$
\State Let $\ProdNode$ be a child for each $\Node \in \Nodes_{\region_1 \cup \region_2}$
\EndFor
\EndFor
\State \Return SPN
\EndProcedure
\end{algorithmic}
\end{algorithm}

\begin{algorithm}[t]
\caption{Random Region Graph}   \label{alg:construct_random_region_graph}
\begin{algorithmic}[1]
\Procedure{RandomRegionGraph}{$\X,D,R$}
\State Create an empty region graph $\rg$
\State Insert $\X$ in $\rg$
\For{$r = 1 \dots R$} 
\State \textproc{Split}$(\rg, \X, D)$
\EndFor
\EndProcedure
\end{algorithmic}
%
%\vspace{\baselineskip}
%
\begin{algorithmic}[1]
\Procedure{Split}{$\rg,\region,D$}
\State Draw balanced partition $\partition = \{\region_1, \region_2\}$ of $\region$
\State Insert $\region_1, \region_2$ in $\rg$
\State Insert $\partition$ in $\rg$
\If{$D>1$}
\If{$|\region_1| > 1$}
%\State 
\textproc{Split}$(\rg, \region_1, D-1)$
\EndIf
\If{$|\region_2| > 1$}
%\State 
\textproc{Split}$(\rg, \region_2, D-1)$
\EndIf
\EndIf
\EndProcedure
\end{algorithmic}
\end{algorithm}
In this paper we construct random regions graphs, with the simple procedure depicted in Algorithm ~\ref{alg:construct_random_region_graph}. 
We randomly divide the root region into two sub-regions of equal size and proceed recursively down to depth $D$, resulting in an SPN of depth $2\,D$.
This recursive splitting mechanism is repeated $R$ times. 
Figure~\ref{fig:example_RAT-SPN} shows an example SPN with $C=3$, $S=2$, and $I=3$, following Algorithm~\ref{alg:construct_random_region_graph} and subsequently Algorithm~\ref{alg:construct_spn_from_region_graph}.
Note that this construction scheme yields SPNs where input distributions, sums, and products can be naturally organized in alternating layers.
Similar to classical multilayer perceptrons (MLPs), each layer takes inputs from its directly preceding layer only. 
Unlike MLPs, however, layers in RAT-SPNs are connected block-wise sparsely in a random fashion.
Thus, layers in MLPs and RAT-SPNs are hardly comparable;
however, we suggest to understand each pair of sum and product layer to be roughly corresponding to one layer in an MLP: sum layers play the role of (block-wise sparse) matrix multiplication and product layers as non-linearities (or, more precisely, bi-linearities of their inputs).
The input layer, containing the SPN's leaf distributions, can be interpreted as a non-linear feature extractors.
%Thus, SPNs can be interpreted as a relative of radial basis networks \cite{Broomhead1988}.

\subsection{TRAINING AND IMPLEMENTATION}

Let $\mathcal{X} = \{(\x_1,y_1), \dots, (\x_N,y_N)\}$ be a training set of inputs $\x_n$ and class labels $y_n$.
Furthermore, let $\SPN_{c}$ be the $c^{\text{th}}$ output of the RAT-SPN and $\bm{\w}$ all SPN parameters.
We train RAT-SPNs by minimizing the objective 
\begin{equation}   \label{eq:rat_objective}
 \mathsf{O}(\bm{\w}) = \lambda \, \mathsf{CE}(\bm{\w}) + (1 - \lambda) \, \mathsf{nLL}(\bm{\w}),
\end{equation}
where $\mathsf{CE}(\bm{\w})$ is the cross-entropy 
\begin{equation}
\mathsf{CE}(\bm{\w}) = - \frac{1}{N} \sum_n \log \frac{\SPN_{y_n}(\x_n)}{\sum_{y'} \SPN_{y'}(\x_n)}
\end{equation}
and $\mathsf{nLL}(\bm{\w})$ is the normalized negative log-likelihood 
\begin{equation}
\mathsf{nLL}(\bm{\w}) = - \frac{1}{N\,|\X|} \sum_n \log \SPN_{y_n}.
\end{equation}
When setting $\lambda=1$, we purely optimize cross-entropy (discriminative setting), while for $\lambda=0$ we perform maximum likelihood training (generative setting).
For $0 < \lambda < 1$, we have a continuum of hybrid objectives, trading off the generative and discriminative character of the model.

We implemented RAT-SPNs in Python/TensorFlow, where the nodes of a region are represented by a matrix with rows corresponding to samples in a mini-batch (containing $100$ samples throughout our experiments) and columns corresponding to the number of distributions in the region (either $I$, $S$ or $C$).
All computation are performed in the log-domain, using the well known \emph{log-sum-exp} trick, readily provided in Tensorflow.
Sum-weights, which we require to be non-negative and normalized, are re-parameterized via log-softmax layers.
Product tensors are implemented by taking outer products (actually sums in the log-domain) of the two matrices below, realized by broadcasting.
Throughout our experiments, we used Adam \cite{Kingma2015} in its default settings.
As input distributions, we used Gaussian distributions with isotropic covariances, i.e.~each input distribution further decomposes into a product of single dimensional Gaussians.
We tried to optimize the variances jointly with the means which, however, delivered worse results than merely setting all variances uniformly to $1$.
While RAT-SPNs are implemented and trained in a seemingless way, they still yield hundreds of tensors.
This, together with performing computations in the log-domain, causes that RAT-SPNs are an order of magnitude slower than ReLU-MLPs of similar sizes.
This disadvantage is mainly an effect of the simplicity of our implementation, just employing native Tensorflow operations and a few dozens of python code.
The advantage of this implementation, however, is that combinations of RAT-SPNs with other deep learning methods can be done in a simple plug and play manner.
Moreover, we are confident that with sufficient engineering effort RAT-SPNs could be trained and run at comparable speed as MLPs.

\subsection{PROBABILISTIC DROPOUT}
\label{sec:dropout}

The size of RAT-SPNs can be easily controlled via the structural parameters $D$, $R$, $S$ and $I$.
RAT-SPNs with many parameters, however, tend to overfit just like regular neural networks, which requires regularization.
One of the classical techniques that boosted deep learning models is the well-known \emph{dropout} heuristic~\cite{Srivastava2014}, setting inputs and/or hidden units to zero with a certain probability $p$, and rescaling the remaining units by $\frac{1}{p}$.
In the following we modify the dropout heuristic for RAT-SPNs, exploiting their probabilistic nature.

\subsubsection{Dropout at Inputs: Marginalizing out Inputs}

Dropout at inputs essentially marks input features as missing at random.
In the probabilistic paradigm, we would simply marginalize over these missing features.
Fortunately, this is an easy exercise in SPNs, as we only need to set the distributions corresponding to the dropped-out features to $1$.
As we operate in the log-domain, this means to set the corresponding log-distribution nodes to $0$.
This is in fact quite similar to standard dropout, except that we are not compensating by $\frac{1}{p}$, and blocks of units are dropped out (i.e., all log-distributions whose scope corresponds to a missing input feature are jointly set to $0$).
%; we learn with form of blankout noise akin to \cite{maatenCTW13}.

\subsubsection{Dropout at Sums: Injection of Discrete Noise}

As discussed in \cite{Poon2011,Zhao2015,Peharz2016}, sum nodes in SPNs can be interpreted as marginalized latent variables, akin to the latent variable interpretation in mixture models.
In particular, \cite{Peharz2016} introduced so-called \emph{augmented} SPNs which explicitly incorporate these latent variables in the SPN structure.
The augmentation introduces indicator nodes representing the states of the latent variables, which can switch the children of sum nodes on or off by connecting them via an additional product.
This mechanism establishes the explicit interpretation of sum children as conditional distributions.

In RAT-SPNs, we can equally well interpret a whole \emph{region} as a single latent variable, and the weights of each sum node in this region as the conditional distribution of this variable.
Indeed, the argumentation in \cite{Peharz2016} also holds when introducing a set of indicators for a single latent variable which is shared by all sum nodes in one region.
While the latent variables are not observed, we can employ a simple probabilistic version of dropout, by introducing artificial observations for them.
For example, if the sum nodes in a particular region have $K$ children (i.e.~the corresponding variable $Z$ has $K$ states), then we could introduce artificial information that $Z$ assumes a state in some \emph{subset} of $\{1,\dots,K\}$.
By doing this for each latent variable in the network, we essentially select a small sub-structure of the whole SPN to explain the data -- this argument is very similar to the original dropout proposal \cite{Srivastava2014}.
Implementing dropout at sum-layers is again straightforward: we select a subset of all product nodes which are connected to the sums in one region and set them to 0 (actually $-\infty$ in the log-domain).
% Again we do not need to multiply with a correction factor.

%%%%%%%%%%%%%%%%%%%%%%%%%%%%%%%%%%%%%%%%%%%%%%%%%%%%%%%%%%%%%%%%%%%%%%%%%%%%%%%%%%%%%%
%%%%%%%%%%%%%%%%%%%%%%%%%%%%%%%%%%%%%%%%%%%%%%%%%%%%%%%%%%%%%%%%%%%%%%%%%%%%%%%%%%%%%%
%%%%%%%%%%%%%%%%%%%%%%%%%%%%%%%%%%%%%%%%%%%%%%%%%%%%%%%%%%%%%%%%%%%%%%%%%%%%%%%%%%%%%%
\section{EXPERIMENTS}   \label{sec:experiments}

\subsection{EXPLORING THE CAPACITY OF RAT-SPNS}   \label{sec:capacity}

In our first experiment, we aim to empirically investigate the capacity of RAT-SPNs, by simply trying to overfit data with various model sizes. 
To this end, we fit RAT-SPNs on MNIST train data, using every combination of split depth $D \in \{1,2,3,4\}$, number of split repetitions $R \in \{10,20,40,80\}$ and number of distributions per region $S=I \in \{5,10,20,40\}$.
In this paper, we follow a data-agnostic setting, i.e.~we deliberately do not exploit the neighborhood  correlations present in images. 
Consequently, our models will perform the same for any permutation of pixels.
The natural baselines in the data-agnostic setting are MLPs, where we take ReLU activations for the hidden units and linear activations for the output layer.
We ran MLPs with every combination of number of layers in $\{1,2,3,4\}$ and number of hidden units in $\{100, 250, 500, 1000, 2000\}$.
For both RAT-SPNs and MLPs, we used Adam with its default parameters to optimize cross-entropy.
\begin{figure}
\begin{centering}
\includegraphics[width=0.995\columnwidth]{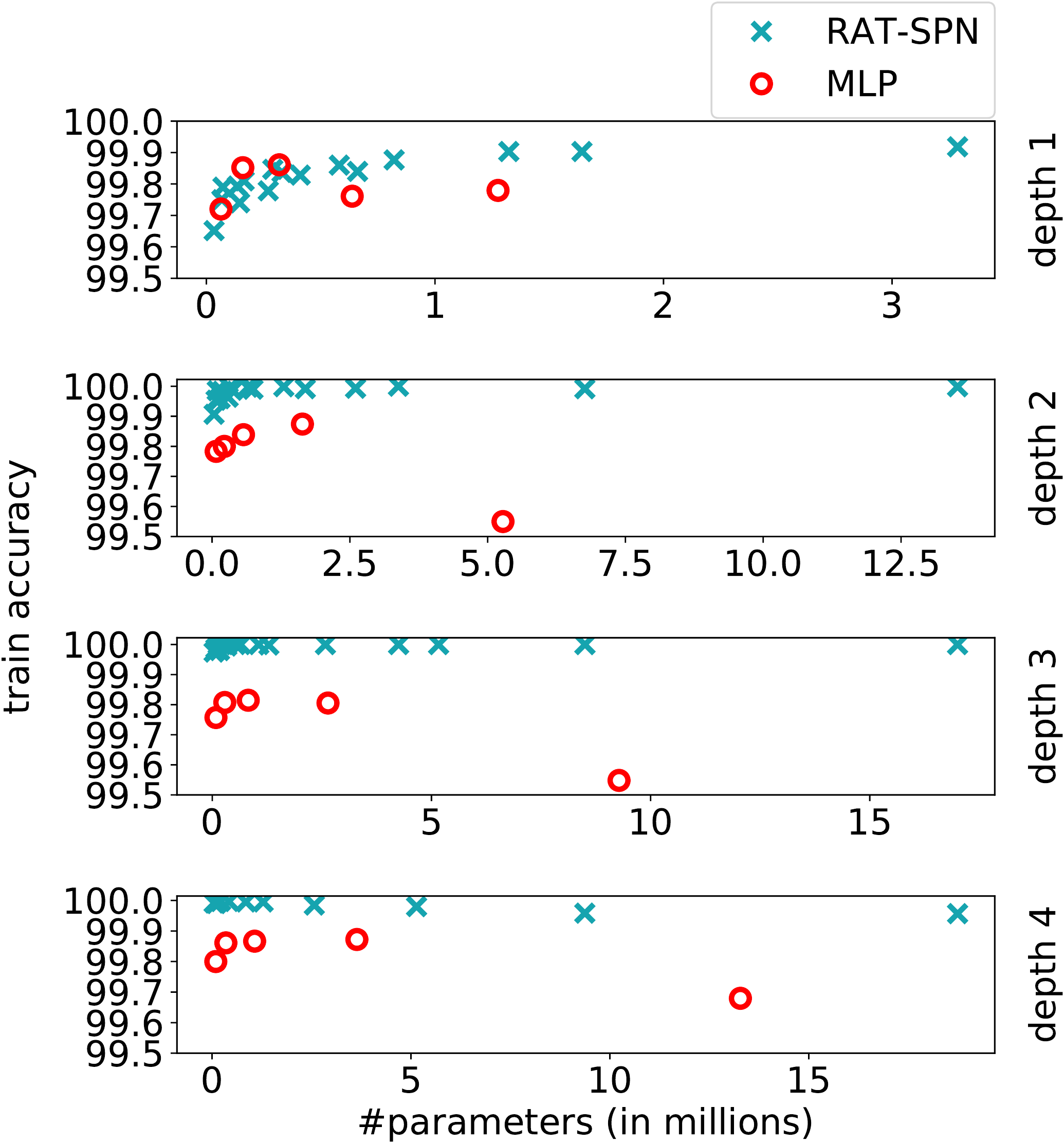}
\caption{
Capacity study by overfitting on MNIST for RAT-SPNs and MLPs (ReLU). 
The y-axis the training accuracy after 200 epochs, as a function of the number of parameters. 
The 'depth' refers to the number of hidden layers in MLP and to split depth $D$ in RAT-SPNs. (Best viewed in color).
}   \label{fig:comparison_overfit}
\end{centering}
\end{figure}
Figure \ref{fig:comparison_overfit} summarizes the training accuracy of both models after 200 epochs as a function of the number of parameters in the respective model.
As one can see see, RAT-SPNs can scale to millions of parameters, and furthermore, they are easily able to overfit the MNIST training set to the same extent as MLPs.
For numbers of layers $2,3,4$ it seems that RAT-SPNs are suited slightly better to fit the data.
This is in fact an artifact of SGD optimization: MLPs still jitter around $100\%$ during the last epochs, while the accuracy of RAT-SPNs remains stable.

These overfitting results give evidence that RAT-SPNs are capacity-wise at least as powerful as ReLU-MLPs. 
In the next experiment, we investigated whether RAT-SPNs are also on par with MLPs concerning generalization on classification tasks.
Subsequently, we show that RAT-SPNs exhibit superior performance when dealing with missing features and are able to identify outliers reliably.

\subsection{GENERALIZATION OF RAT-SPNS}

When trained without regularization, RAT-SPNs achieve less than $97\%$ on the test set of MNIST, which is rather inferior even for data-agnostic models.
Therefore, we trained them with our probabilistic dropout variant as introduced in section~\ref{sec:dropout}.
We cross-validated $D \in \{2,3\}$, $R \in \{20,40\}$ and number of distributions per region $S=I \in \{10,20,30,40\}$, dropout rates for inputs in $\{1.0, 0.75, 0.5, 0.25\}$ and dropout rates for sum-layers in $\{1.0, 0.75, 0.5, 0.25\}$.
A dropout rate of $p$ means that a fraction of $p$ features is kept on average.

For comparison, we trained ReLU-MLPs with number of hidden layers in $\{1,2,3,4\}$, number of hidden units in $\{100, 250, 500, 1000, 2000\}$, input dropout rates in $\{1.0, 0.75, 0.5, 0.25\}$ and dropout rates for hidden layers in $\{1.0, 0.75, 0.5, 0.25\}$. 
No dropout was applied to the output layer.
We trained MLPs in two variants, namely 'vanilla' (vMLPs), meaning that besides dropout no additional optimization tricks were applied, and a variant (MLP) also employing Xavier-initialization \cite{Glorot2010} and batch normalization \cite{Ioffe2015}.
While the latter should be considered the default variant to train MLPs, note that helpful heuristics like Xavier-initialization and batch normalization have evolved over decades, while similar techniques for RAT-SPNs are not available.
Thus, vMLPs might serve as a fairer comparison.

\begin{table}
\begin{center}
\begin{tabular}{lllll}
\toprule
       & &   \multicolumn{1}{c}{\bf RAT-SPN}    &   \multicolumn{1}{c}{\bf MLP}   &   \multicolumn{1}{c}{\bf vMLP}     \\
\midrule
\multirow{6}{*}{\rotatebox[origin=c]{90}{Accuracy}} &   MNIST           &    98.19  &   98.32   &   98.09   \\
&               &   (8.5M)  &   (2.64M) &   (5.28M) \\
&   F-MNIST     &   89.52   &   90.81   &   89.81   \\
&               &   (0.65M) &   (9.28M) &   (1.07M) \\
&   20-NG       &   47.8    &   49.05   &   48.81   \\
&               &   (0.37M) &   (0.31M) &   (0.16M) \\
\midrule 
\multirow{6}{*}{\rotatebox[origin=c]{90}{Cross-Entropy}} &   MNIST          &    0.0852 &   0.0874  &   0.0974   \\
&               &   (17M)   &   (0.82M) &   (0.22M) \\
&   F-MNIST     &   0.3525  &   0.2965  &   0.325   \\
&               &   (0.65M) &   (0.82M) &   (0.29M) \\
&   20-NG       &   1.6954  &   1.6180  &   1.6263  \\
&               &   (1.63M) &   (0.22M) &   (0.22M) \\
\bottomrule
\end{tabular}
\end{center}
\caption{
Classification results for MNIST, fashion MNIST (F-MNIST) and 20 News Groups (20-NG) using RAT-SPNs, MLPs and 'Vanilla MLPs' (vMLP, trained without Xavier-initialization and batch normalization).
Best test values for accuracy and Cross-Entropy are reported, as well as the corresponding number of parameters in the model (in parenthesis).}
\label{tab:classification_results}
\end{table}

We trained on MNIST, fashion-MNIST\footnote{Fashion-MNIST is a dataset in the same format as MNIST, but with the task of classifying fashion items rather than digits; github.com/zalandoresearch/fashion-mnist} and 20 News Groups  (20-NG). % \footnote{scikit-learn.org/stable/datasets/twenty\_newsgroups.html}
The 20-NG dataset is a text corpus of 18846 news documents that belong to 20 different news groups or classes. 
We first split the news documents into 13568 instances for training, 1508 for validation, and 3770 for testing. 
The text was pre-processed into a bag-of-words representation by keeping the top 1000 most relevant words according to their Tf-IDF. 
Then, 50 topics were extracted using LDA \cite{Blei2003} and employed as the new feature representation for classification.

Table \ref{tab:classification_results} summarizes the classification accuracy and cross-entropy on the test set, as well as the size of the models in terms of number of parameters. 
As one can see, RAT-SPNs are on par with MLPs, and only slightly outperformed in terms of traditional classification tasks.
However, as shown in the following sections, the real potential of probabilistic deep learning models actually lies beyond classical benchmark results.

\subsection{HYBRID POST-TRAINING}

\begin{figure}[t]
\begin{centering}
\includegraphics[width=0.995\columnwidth]{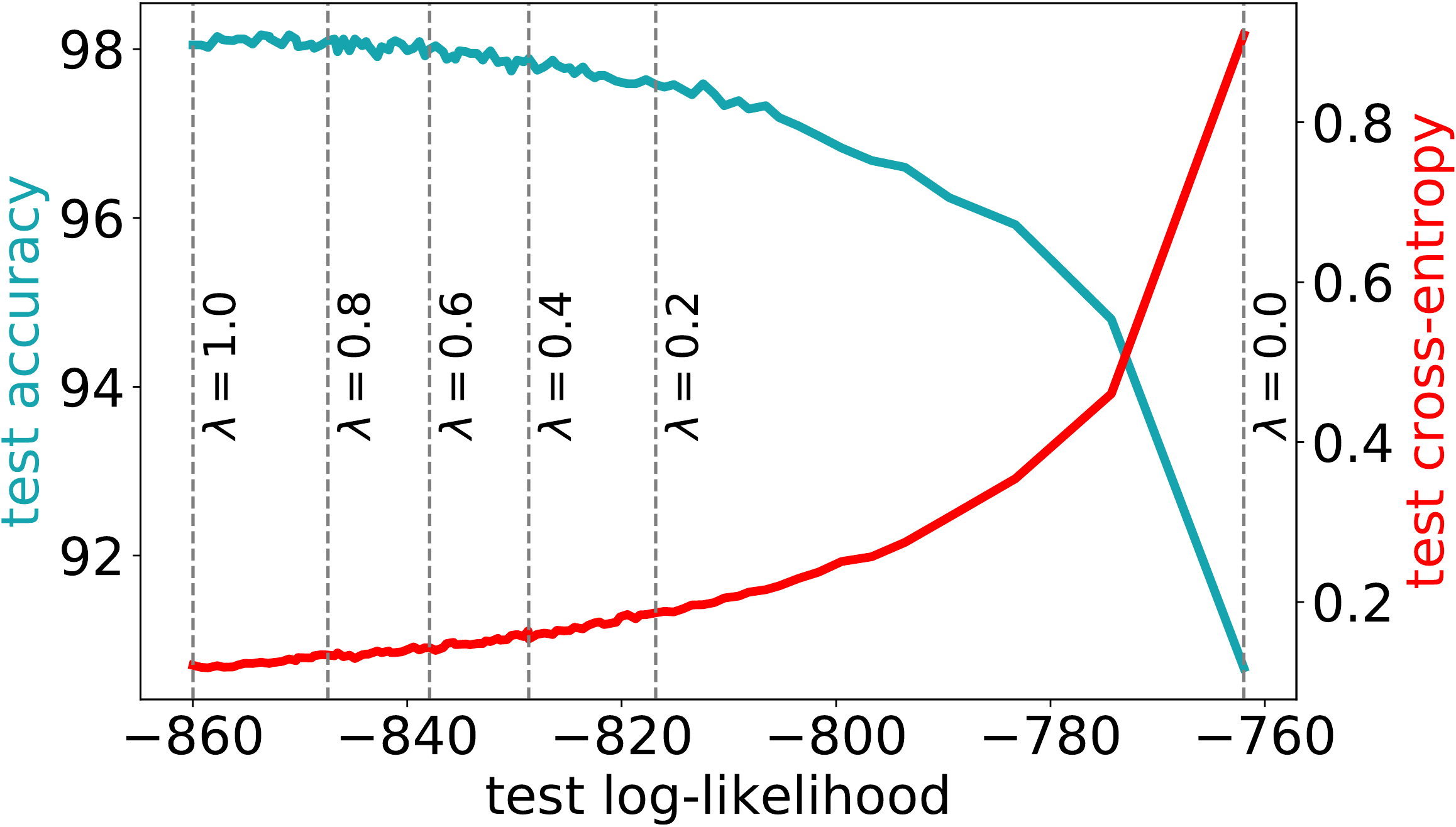}
\caption{
A RAT-SPN is a joint model over both inputs and classes and allows to evaluate the likelihood over the inputs.
By varying $\lambda$ we can control the trade-off between generative behavior (measured in log-likelihood) and discriminative behavior (measured in accuracy or cross-entropy).
}   \label{fig:hybrid_posttrain}
\end{centering}
\end{figure}

Recall that SPNs define a full joint distribution over both inputs $\X$ and class variable $C$, and that our objective \eqref{eq:rat_objective} with trade-off parameter $\lambda$ allows us to trade off between cross-entropy ($\lambda=1$) and log-likelihood ($\lambda=0$).
When $\lambda=1$, we cannot hope that the distribution over $\X$ is faithful to the underlying data.
By setting $\lambda < 1$, however, we can obtain interesting hybrid models, yielding both a discriminative and generative behavior. 
To this end, we use the RAT-SPN with highest validation accuracy from the previous experiment, and post-train it for another 20 epochs, for various values of $0 \leq \lambda \leq 1$. 
This yields a natural trade-off between the log-likelihood over inputs $\X$ and predictive performance regarding classification accuracy/cross-entropy.
Figure~\ref{fig:hybrid_posttrain} shows this trade-off.
As one can see, by sacrificing little predictive performance, we can drastically improve the generative character of SPNs.
The benefit of this is shown in the following.

\subsection{SPNS ARE ROBUST UNDER MISSING FEATURES}

\begin{figure}[t]
\begin{centering}
\includegraphics[width=0.9\columnwidth]{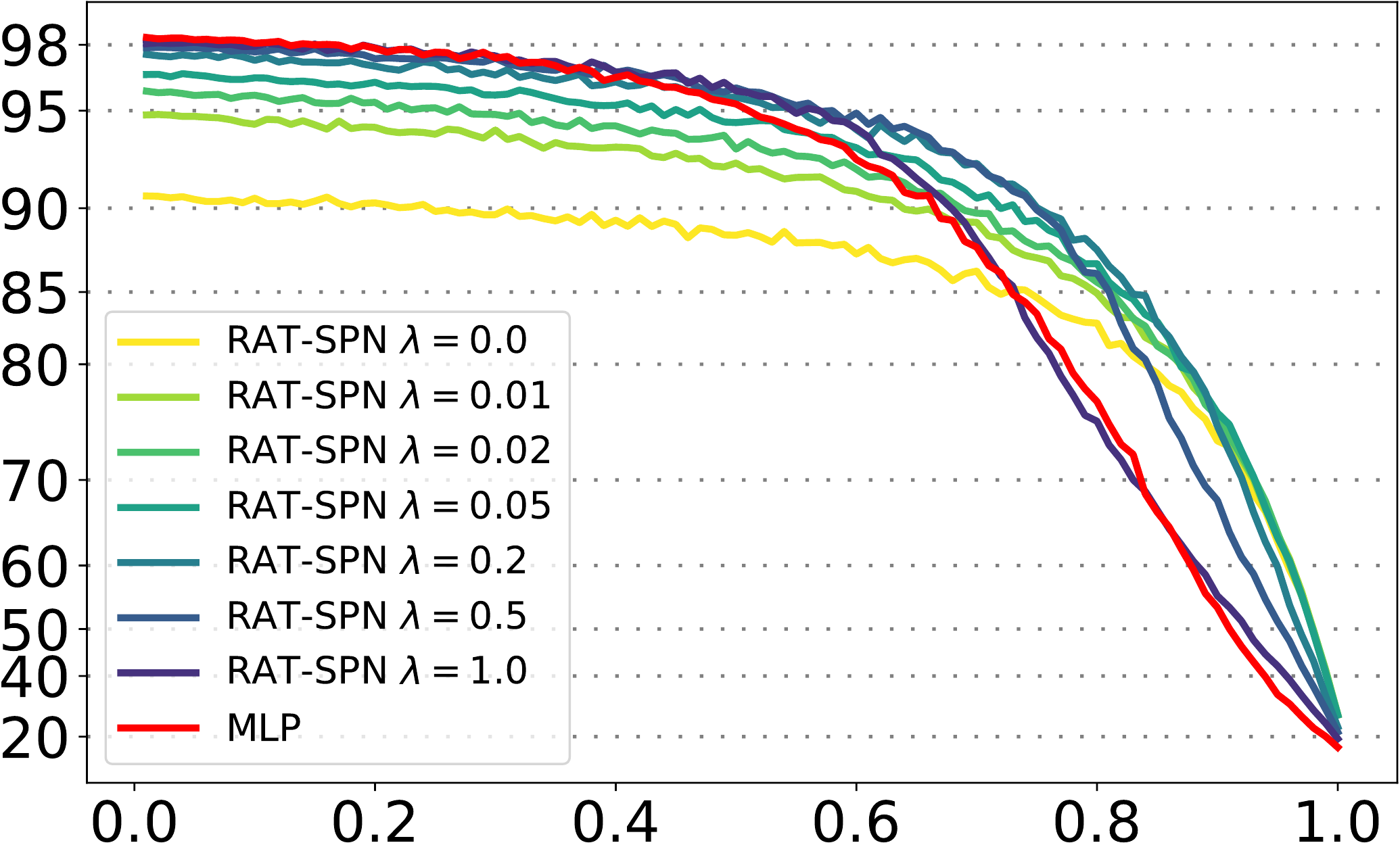}
\caption{
Classification accuracy [\%] of hybrid RAT-SPNs and MLP as a function of percentage $p$ of missing features, varying from $0.0$ (no features missing) to $0.99$ (99\% of features missing).
}   \label{fig:miss-robust}
\end{centering}
\end{figure}

\begin{figure*}
\begin{centering}
\includegraphics[width=0.99\textwidth]{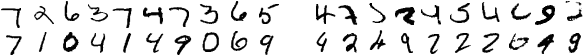}\\%
\vspace{1\baselineskip}
\includegraphics[width=0.99\textwidth]{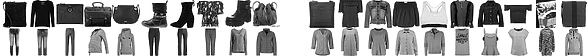}
\caption{
Outliers (samples with log-likelihood $<-2000$) and inliers (samples with log-likelihood $>-850$) on MNIST (top) and fashion-MNIST (bottom) for RAT-SPN post-trained with $\lambda=0.2$.
Samples on the left half were classified correctly, samples on the right half were classified incorrectly.
The upper rows are outliers, the lower rows are inliers, for MNIST and fashion-MNIST, respectively.
The predictions for wrong MNIST digits are (depicted as correct$\rightarrow$predicted): (top-row) 4$\rightarrow$2, 7$\rightarrow$3, 5$\rightarrow$3, 2$\rightarrow$9, 4$\rightarrow$7, 5$\rightarrow$3, 4$\rightarrow$6, 6$\rightarrow$2, 9$\rightarrow$3, 2$\rightarrow$6;
(bottom-row) 4$\rightarrow$9, 2$\rightarrow$8, 4$\rightarrow$2, 9$\rightarrow$4, 2$\rightarrow$7, 2$\rightarrow$7, 2$\rightarrow$8, 6$\rightarrow$0, 4$\rightarrow$9, 9$\rightarrow$8.
}   \label{fig:outlier_inlier}
\end{centering}
\end{figure*}

When input features in $\X$ are missing at random, the probabilistic paradigm dictates to marginalize these \cite{Little2014}.
As SPNs allow marginalization simply and efficiently, we expect that RAT-SPNs should be able to robustly treat missing features, especially the ``more generative'' they are (corresponding to smaller $\lambda$).
To this end, we randomly discard a fraction of $p$ pixels in the MNIST test data -- independently for each sample -- and classify the data using RAT-SPNs trained with various values of $\lambda$, marginalizing missing features.
This is the same procedure we used for probabilistic dropout during training, cf.~section~\ref{sec:dropout}.
Similarly, we might expect MLPs to perform robustly under missing features during test time, by applying (classical) dropout.

Figure~\ref{fig:miss-robust} summarizes the classification results when varying $p$ between $0.0$ and $0.99$.
As one can see, RAT-SPNs with smaller $\lambda$ are more stable under even large fractions of missing features.
A particularly interesting choice is $\lambda=0.2$: here the corresponding RAT-SPN starts with an accuracy $97.58\%$ for no missing features and degrades very gracefully: for a large fraction of missing features ($>60\%$) the advantage over MLPs is dramatic.
Note that this result is consistent with other hybrid learning schemes applied in graphical models \cite{Peharz2013b}.
Purely discriminative RAT-SPNs and MLPs are roughly on par concerning robustness against missing features.

\subsection{SPNS KNOW WHAT THEY DON'T KNOW}

Besides being robust against missing features, an important feature of (hybrid) generative models is that they are naturally able to detect outliers and peculiarities by monitoring the likelihood over inputs $\X$.
To this end, we evaluated the likelihoods on the test set for both MNIST and fashion-MNIST, using the respective RAT-SPN post-trained with $\lambda=0.2$.
We selected two thresholds of $-2000$ and $-850$ by visual inspection of the likelihood histograms.
These two values determine roughly the $5\%$ percentiles of most likely/unlikely samples.
In both these sets, we selected -- following the original order in MNIST -- the first 10 samples which are correctly and incorrectly classified, respectively.
Thus yields 4 groups of 10 samples each: outlier/correct, outlier/incorrect, inlier/correct, inlier/incorrect.

These samples are shown in Figure~\ref{fig:outlier_inlier}.
Albeit qualitative, these results are interesting: 
One can visually confirm that the outlier MNIST digits are indeed peculiar, both the correctly and the incorrectly classified ones.
Among the outlier/incorrect group are 2 digits (top row, right, 3rd and 8th), which are not recognizable to the authors either.
The inlier/incorrect digits can be interpreted, to a certain extent, as the ambiguous ones, e.g.~two '2's (bottom row, right, 5th and 6th) are similar to '7' (and indeed classified as such), or a digit (bottom row, right, 8th) which could either be '6' or '0'.
For fashion-MNIST, one can clearly see that the outliers are all low in contrast and fill the whole image.
In one images (top row, right, 9th) the background has not been removed.

For a more objective analysis, we use a variant of \emph{transfer testing} recently proposed by Bradshaw et al.~\cite{Bradshaw2017}.
This technique is quite simple: we feed a classifier trained on one domain (e.g.~MNIST) with examples from a related but different domain, e.g.~street view house numbers (SVHN) \cite{Netzer2011} or the handwritten digits of SEMEION \cite{Buscema1998}, converted to MNIST format ($28 \times 28$ pixels, grey scale).
While we would expect that most classifiers perform poorly in such setting, an important property of an AI system would be to be aware that it is confronted with out-of-domain data and be able to communicate this either to other parts of the system or a human user.
While Bradshaw et al.~applied transfer testing to conditional models in order to assess output uncertainties, we follow an arguably natural approach and assess \emph{input uncertainties} in RAT-SPNs, i.e.~their likelihoods over $\X$.

\begin{figure}
\begin{centering}
\includegraphics[width=0.995\columnwidth]{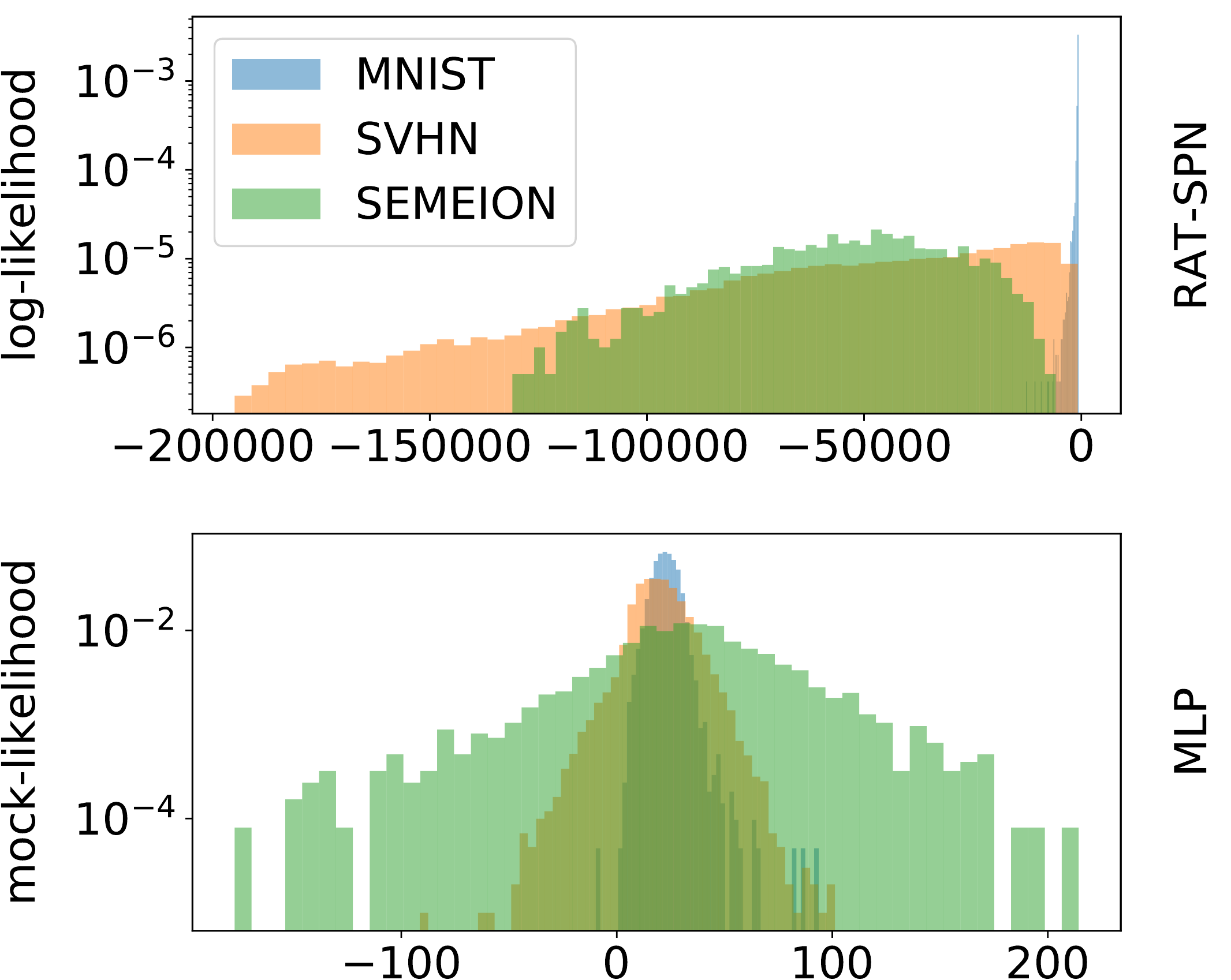}
\caption{
Histograms of test log-likelihoods for MNIST, SVHN and SEMEION data for RAT-SPN (top) and corresponding computations performed for MLP (``mock-likelihood'') (bottom).
Both models were trained on MNIST. 
The likelihoods of RAT-SPNs yield a strong signal whether a sample is in-domain or out-of-domain, while the ``mock-likelihood'' does not discriminate well between these cases.
}   \label{fig:px_histograms}
\end{centering}
\end{figure}

Figure~\ref{fig:px_histograms}, top, shows histograms of the log-likelihoods of the RAT-SPN post-trained with $\lambda=0.2$, when fed with MNIST test data (in-domain), SVHN test data (out-of-domain) and SEMEION (out-of-domain).
The result is striking: the histogram shows that the likelihood over \emph{inputs} provides a strong signal (note the y-axis log-scale) whether a sample comes from in-domain or out-of-domain.
That is, RAT-SPNs have an additional communication channel to inform us whether we ought trust their predictions.
An MLP does not have such a mean, as it does not represent a full joint distribution.
However, a potential objection could be that this positive results for SPNs (or more generally, joint models) might actually stem in some way from the discriminative character of the model, rather than from its generative nature.
After all, in order to compute SPN likelihoods in Figure~\ref{fig:px_histograms}, we simply had to sum over the SPN outputs, re-weighted by a class prior (assumed uniform here).
Perhaps the strong outlier-detection signal merely stems from averaging predictive outputs?
Thus, as a sanity check we perform the likewise computations in our trained MLPs.
One might suspect, that the result, although not interpretable as log-probability, still yields a decent signal to detect outliers.
In need of a name for this rather odd quantity, we name it \emph{mock-likelihood}.
Figure~\ref{fig:px_histograms}, bottom, shows histograms of this mock-likelihood: although histograms are more spread for out-of-domain data, they are highly overlapping, 
yielding no clear signal for out-of-domain vs. in-domain.

\section{CONCLUSION}   \label{sec:conclusion}

We introduced a particularly simple but effective way to train SPNs: simply pick a random structure and train them in end-to-end fashion like neural networks.
This makes the application of SPNs within the deep learning framework seamless and allows the application of common deep learning tools such automatic differentiation and easy use of GPUs.
As a modest technical contribution, we adapted the well-known dropout heuristic and equipped it with a sound probabilistic interpretation within RAT-SPNs.
RAT-SPNs show classification performance on par with traditional neural networks on several classification tasks.
Moreover, RAT-SPNs demonstrate their full power when used as a generative model, showing remarkable robustness against missing features through exact and efficient inference and compelling results in anomaly/out-of-domain detection.
In future work, the hybrid properties of RAT-SPNs could allow promising directions like new variants of semi-supervised or active learning.
While this paper is held in the data-agnostic regime, in future we will investigate SPNs tailored to structured data sources.

% \bibliographystyle{plain}
% \bibliography{bibliography}

\begin{thebibliography}{10}

\bibitem{Adel2015}
T.~Adel, D.~Balduzzi, and A.~Ghodsi.
\newblock Learning the structure of sum-product networks via an {SVD}-based
  algorithm.
\newblock In {\em UAI}, 2015.

\bibitem{Blei2003}
D.~M. Blei, A.~Y. Ng, and M.~I. Jordan.
\newblock Latent dirichlet allocation.
\newblock {\em J. Mach. Learn. Res.}, 3, 2003.

\bibitem{Bradshaw2017}
J.~Bradshaw, A.~Matthews, and Z.~Ghahramani.
\newblock Adversarial examples, uncertainty, and transfer testing robustness in
  gaussian process hybrid deep networks.
\newblock {\em preprint arXiv}, 2017.
\newblock arxiv.org/abs/1707.02476.

\bibitem{Buscema1998}
M.~Buscema.
\newblock {\em MetaNet*: The Theory of Independent Judges}, volume~33.
\newblock 02 1998.

\bibitem{Darwiche2003}
A.~Darwiche.
\newblock A differential approach to inference in {B}ayesian networks.
\newblock {\em Journal of the ACM}, 50(3):280--305, 2003.

\bibitem{Dennis2012}
A.~Dennis and D.~Ventura.
\newblock Learning the architecture of sum-product networks using clustering on
  variables.
\newblock In {\em Proceedings of NIPS}, 2012.

\bibitem{Abadi2015}
Abadi~M. et~al. (40~authors).
\newblock {TensorFlow}: Large-scale machine learning on heterogeneous systems,
  2015.

\bibitem{Gens2012}
R.~Gens and P.~Domingos.
\newblock Discriminative learning of sum-product networks.
\newblock In {\em Proceedings of NIPS}, pages 3248--3256, 2012.

\bibitem{Gens2013}
R.~Gens and P.~Domingos.
\newblock Learning the structure of sum-product networks.
\newblock {\em Proceedings of ICML}, pages 873--880, 2013.

\bibitem{Glorot2010}
X.~Glorot and Y.~Bengio.
\newblock Understanding the difficulty of training deep feedforward neural
  networks.
\newblock In {\em Proceedings of AISTATS}, pages 249--256, 2010.

\bibitem{Goodfellow2014}
I.~J. Goodfellow, J.~Pouget-Abadie, M.~Mirza, B.~Xu, D.~Warde-Farley, S.~Ozair,
  A.~Courville, and Y.~Bengio.
\newblock Generative adversarial nets.
\newblock In {\em Proceedings of NIPS}, pages 2672--2680, 2014.

\bibitem{Ioffe2015}
S.~Ioffe and C.~Szegedy.
\newblock Batch normalization: Accelerating deep network training by reducing
  internal covariate shift.
\newblock In {\em Prooceedings of ICML}, 2015.

\bibitem{Kingma2015}
D.~P. Kingma and J.~Ba.
\newblock Adam: A method for stochastic optimization.
\newblock In {\em Proceedings of ICLR}, 2015.

\bibitem{Kingma2014}
D.~P. Kingma and M.~Welling.
\newblock Auto-encoding variational {Bayes}.
\newblock In {\em ICLR}, 2014.
\newblock arXiv:1312.6114.

\bibitem{Larochelle2011}
H.~Larochelle and I.~Murray.
\newblock The neural autoregressive distribution estimator.
\newblock In {\em Proceedings of AISTATS}, pages 29--37, 2011.

\bibitem{Little2014}
R.~J.~A. Little and D.~B Rubin.
\newblock {\em Statistical analysis with missing data}, volume 333.
\newblock John Wiley \& Sons, 2014.

\bibitem{Martens2015}
J.~Martens and V.~Medabalimi.
\newblock On the expressive efficiency of sum product networks.
\newblock online, http://arxiv.org/abs/1411.7717, 2015.

\bibitem{Molina2018}
A.~Molina, A.~Vergari, N.~{Di Mauro}, S.~Natarajan, F.~Esposito, and
  K.~Kersting.
\newblock Mixed sum-product networks: A deep architecture for hybrid domains.
\newblock In {\em Proceedings of AAAI}, 2018.

\bibitem{Netzer2011}
Y.~Netzer, T.~Wang, A.~Coates, A-Bissacco, B.~Wu, and A.~Y. Ng.
\newblock Reading digits in natural images with unsupervised feature learning.
\newblock In {\em NIPS Workshop on Deep Learning and Unsupervised Feature
  Learning 2011}, 2011.

\bibitem{Peharz2013}
R.~Peharz, B.~Geiger, and F.~Pernkopf.
\newblock Greedy part-wise learning of sum-product networks.
\newblock In {\em Proceedings of ECML/PKDD}, pages 612--627. Springer Berlin,
  2013.

\bibitem{Peharz2016}
R.~Peharz, R.~Gens, F.~Pernkopf, and P.~Domingos.
\newblock On the latent variable interpretation in sum-product networks.
\newblock {\em IEEE Transactions on Pattern Analysis and Machine Intelligence},
  2016.

\bibitem{Peharz2014a}
R.~Peharz, G.~Kapeller, P.~Mowlaee, and F.~Pernkopf.
\newblock Modeling speech with sum-product networks: Application to bandwidth
  extension.
\newblock In {\em Proceedings of ICASSP}, pages 3699--3703, 2014.

\bibitem{Peharz2013b}
R.~Peharz, S.~Tschiatschek, and F.~Pernkopf.
\newblock The most generative maximum margin {Bayesian} networks.
\newblock In {\em Proceedings of ICML}, pages 235--243, 2013.

\bibitem{Peharz2015b}
R.~Peharz, S.~Tschiatschek, F.~Pernkopf, and P.~Domingos.
\newblock On theoretical properties of sum-product networks.
\newblock In {\em Proceedings of AISTATS}, pages 744--752, 2015.

\bibitem{Poon2011}
H.~Poon and P.~Domingos.
\newblock Sum-product networks: A new deep architecture.
\newblock In {\em Proceedings of UAI}, pages 337--346, 2011.

\bibitem{Rashwan2016}
A.~Rashwan, H.~Zhao, and P.~Poupart.
\newblock Online and distributed bayesian moment matching for parameter
  learning in sum-product networks.
\newblock In {\em AISTATS}, pages 1469--1477, 2016.

\bibitem{Rezende2014}
D.~J. Rezende, S.~Mohamed, and D.~Wierstra.
\newblock Stochastic backpropagation and approximate inference in deep
  generative models.
\newblock In {\em Proceedings of ICML}, pages 1278--1286, 2014.

\bibitem{Rooshenas2014}
A.~Rooshenas and D.~Lowd.
\newblock {Learning Sum-Product Networks with Direct and Indirect Variable
  Interactions}.
\newblock {\em ICML -- JMLR W\&CP}, 32:710--718, 2014.

\bibitem{Srivastava2014}
N.~Srivastava, G.~Hinton, A.~Krizhevsky, I.~Sutskever, and R.~Salakhutdinov.
\newblock Dropout: A simple way to prevent neural networks from overfitting.
\newblock {\em JMLR}, 15:1929--1958, 2014.

\bibitem{Trapp2017}
M.~Trapp, T.~Madl, R.~Peharz, F.~Pernkopf, and R.~Trappl.
\newblock Safe semi-supervised learning of sum-product networks.
\newblock In {\em Proceedings of UAI}, 2017.

\bibitem{Trapp2016}
M.~Trapp, R.~Peharz, M.~Skowron, T.~Madl, F.~Pernkopf, and R.~Trappl.
\newblock Structure inference in sum-product networks using infinite
  sum-product trees.
\newblock In {\em NIPS Workshop on Practical Bayesian Nonparametrics}, 2016.

\bibitem{Uria2014}
B.~Uria, I.~Murray, and H.~Larochelle.
\newblock A deep and tractable density estimator.
\newblock In {\em Proceedings of ICML}, pages 467--475, 2014.

\bibitem{vandenOord2016}
A.~van~den Oord, N.~Kalchbrenner, and K.~Kavukcuoglu.
\newblock Pixel recurrent neural networks.
\newblock In {\em Proceedings of ICML}, 2016.

\bibitem{Vergari2015}
A.~Vergari, N.~Di~Mauro, and F.~Esposito.
\newblock Simplifying, regularizing and strengthening sum-product network
  structure learning.
\newblock In {\em Proceedings of ECML/PKDD}, pages 343--358. Springer, 2015.

\bibitem{Vergari2018}
A.~Vergari, R.~Peharz, N.~Di~Mauro, A.~Molina, K.~Kersting, and F.~Esposito.
\newblock Sum-product autoencoding: Encoding and decoding representations using
  sum-product networks.
\newblock In {\em AAAI}, 2018.

\bibitem{Zhao2016}
H.~Zhao, T.~Adel, G.~Gordon, and B.~Amos.
\newblock Collapsed variational inference for sum-product networks.
\newblock In {\em Proceedings of ICML}, 2016.

\bibitem{Zhao2015}
H.~Zhao, M.~Melibari, and P.~Poupart.
\newblock On the relationship between sum-product networks and {B}ayesian
  networks.
\newblock In {\em Proceedings of ICML}, 2015.

\bibitem{Zhao2016b}
H.~Zhao, P.~Poupart, and G.~J Gordon.
\newblock A unified approach for learning the parameters of sum-product
  networks.
\newblock In {\em Proceedings of NIPS}. 2016.

\end{thebibliography}

\end{document}